\documentclass[conference]{IEEEtran}
\IEEEoverridecommandlockouts

\usepackage{cite}
\usepackage{amsmath,amssymb,amsfonts}
\usepackage{graphicx}
\usepackage{algorithmic}
\usepackage{algorithm}
\usepackage{textcomp}
\usepackage{xcolor}
\usepackage{comment}
\usepackage{multirow}

\def\BibTeX{{\rm B\kern-.05em{\sc i\kern-.025em b}\kern-.08em
    T\kern-.1667em\lower.7ex\hbox{E}\kern-.125emX}}

\begin{document}

\title{A Machine Learning Approach to Detect Customer Satisfaction From Multiple Tweet Parameters}

\author{\IEEEauthorblockN{Md. Mahmudul Hasan}
\IEEEauthorblockA{\textit{Department of Electrical and Electronics Engineering} \\
\textit{Bangladesh University of Engineering and Technology}\\
Dhaka, Bangladesh \\
0421062325@eee.buet.ac.bd}
\and
\IEEEauthorblockN{Dr. Shaikh Anowarul Fattah}
\IEEEauthorblockA{\textit{Department of Electrical and Electronics Engineering} \\
\textit{Bangladesh University of Engineering and Technology}\\
Dhaka, Bangladesh \\
fattah@eee.buet.ac.bd
}
}

\maketitle

\begin{abstract}
Since internet technologies have advanced, one of the primary factors for company development is customer happiness. Online platforms have become prominent places for sharing reviews. Twitter is one of these platforms where customers frequently post their thoughts. Reviews of flights on these platforms have become a concern for the airline business. A positive review can help the company grow, while a negative one can quickly ruin its revenue and reputation. So it’s vital for airline businesses to examine the feedback and experiences of their customers and enhance their services to remain competitive. But studying thousands of tweets and analyzing them to find the satisfaction of the customer is quite a difficult task. This tedious process can be made easier by using a machine-learning approach to analyze tweets to determine client satisfaction levels. Some work has already been done on this strategy to automate the procedure using machine learning and deep learning techniques. However, they are all purely concerned with assessing the text’s sentiment. In addition to the text, the tweet also includes the time, location, username, airline name, and so on. This additional information can be crucial for improving the model's outcome. To provide a machine learning-based solution, this work has broadened its perspective to include these qualities. And it has come as no surprise that the additional features beyond text sentiment analysis produce better outcomes in machine-learning-based models.
\end{abstract}


\section{Introduction}
With the increase of the mobile phone, social media users are increasing daily with the advancement of technology. The number of smartphone users is currently 6.657 billion worldwide, which will be 7.690 billion at the end of 2027\cite{b1}. Famous social media platforms like Facebook, Twitter, and Instagram can be easily accessed through a smartphone. As a result, anyone can now express their concerns and experiences on these platforms. The users of these platforms share their emotions, sentiments, and other feelings about their experience of a restaurant, a journey, or a vacation trip, which can be positive, negative, or neutral. People are more open on these platforms than the emotions they express during face-to-face communication. So, the experiences shared on these platforms are vital for any service provider. Hence, the airline service providers can improve their service from the feedback people share on social platforms like Twitter and Facebook.\\
Customer: who purchases any service or product\cite{b2}. The customer's feelings about the service or product can be positive, negative, or neutral with respect to the service, or the product's value can be defined as customer satisfaction\cite{b3}. As airlines are substantial companies, many people use their services and share experiences over the platform. It is difficult to track or trace customers' feedback from the text they tweet. Automating this process will be very beneficial for the airlines. Thanks to the development of the technology space, machine learning has come up with a solution for this. This work aims to provide a machine learning-based solution for automating the process of detecting or extracting feedback from customer tweets.\\

\section{Similar Studies}
Natural language processing is still in its infancy but is expanding quickly. There is a lot of literature in the area of natural language processing. The most well-known area of natural language processing is sentiment analysis. Many studies have previously been conducted on the sentiment analysis of text data acquired from various online sites. But there is still a long way to go in terms of precision and effectiveness.\\

A machine-learning-based approach was created by Kumar and Zymbler to categorize emotions from Twitter content\cite{b4}. For their research, they employed a Python script that was made utilizing the Twitter API\cite{b4}. In this work, the Glove dictionary and the n-gram technique were used to extract characteristics from the text\cite{b14,b15}. The convolutional neural network, artificial neural network, and support vector machine were also used to categorize the emotions\cite{b16}. CNN performed better than ANN and SVM classifiers, they discovered.\\

Soujana et al. concluded that deep learning algorithm-based classifiers outperform other classifiers after doing NLP-based research on several datasets\cite{b5}. Furthermore, by adjusting the input and output layers as necessary, this study was able to advance much farther. Baydogan et al. conducted research on the sentiment analysis of tweeter data from airlines\cite{b6}. Natural language processing (NLP) and machine learning-based classifiers were used to extract sentiment from the text in their study.For the churn prediction challenge, Sabbeh et al. used 10 different machine learning-based analytical techniques\cite{b7}. The investigation's findings showed that the random forest and ADA boost had the highest accuracy at 96\%\cite{b7}.\\

Unfortunately, all of the prior research has only attempted to enhance the performance of the model by focusing on the extraction of emotions from tweet content. No studies have taken into account the additional information that may be gleaned from a tweet, such as a user name, airline name, tweeting time, location, etc. This collection of tweets can be categorized by the location, time, and airline name that can be inferred from the tweets, and the majority of passengers on a flight receive similar levels of service. Thus, these parameters are helpful to improve the prediction model. All of the earlier studies, however, have disregarded it.\\

\section{Dataset Description and Analysis}
For this study, a dataset called "Twitter US Airline Sentiment" including 14,640 tweets from travelers who expressed their feelings on Twitter in February 2015 has been used\cite{b8}. The tweet content, location, time, user name, airline name, and other information are all included in this dataset. The six well-known American airlines Virgin America, United, Southwest, Delta, US Airways, and American have been selected for this study's dataset, which includes tweets about them\cite{b8}. However, several columns hardly contain any data and these was removed during preparation. Three distinct sentiment categories have been applied to the dataset: good, neutral, and negative classifications

\vspace{-0.1in}
\begin{figure}[H]
\centerline{\includegraphics[width=0.56\textwidth]{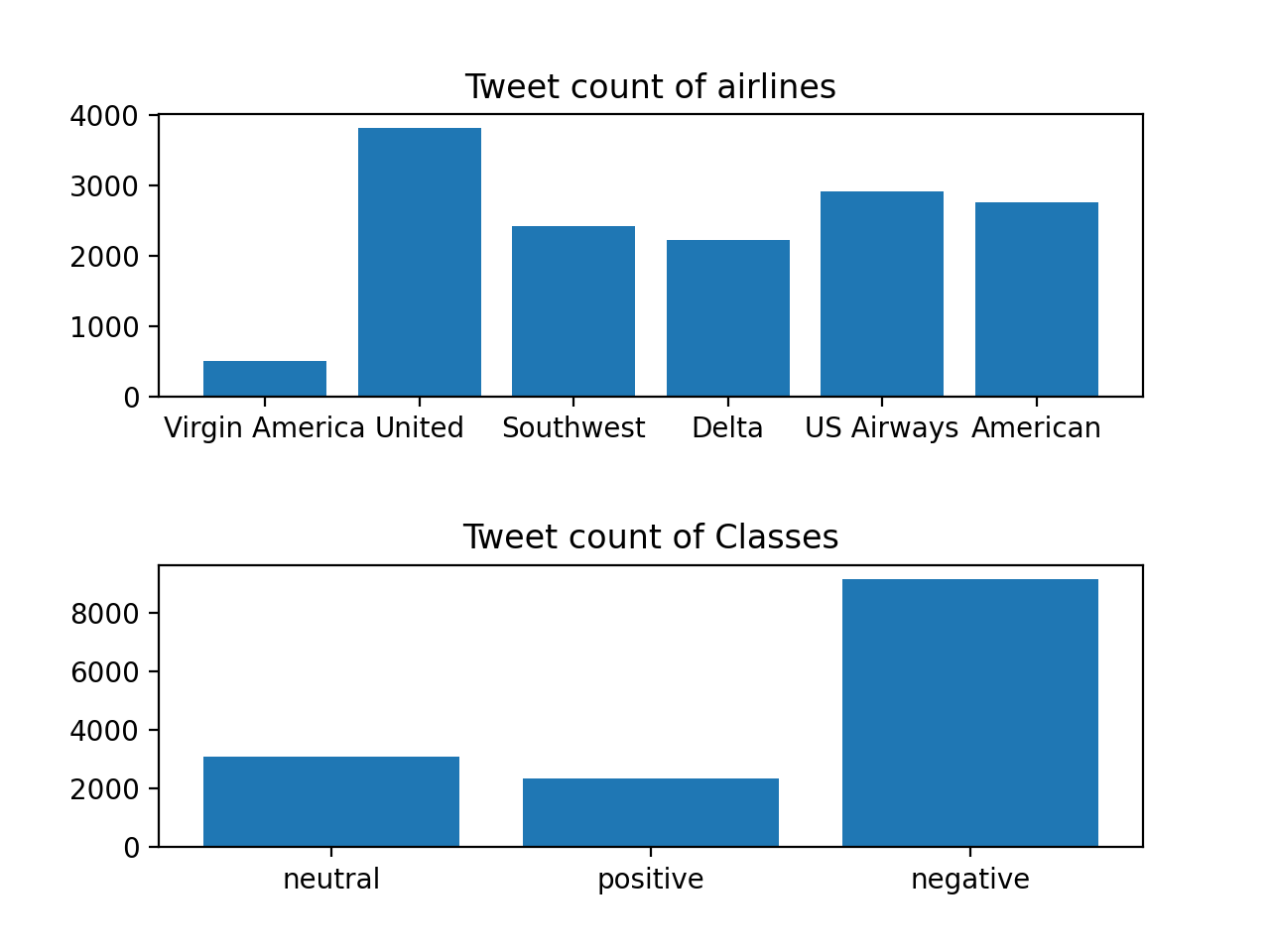}}
\caption{Tweet count of different airlines}
\label{fig1}
\end{figure}

\section{Used Methodologies}
This project aims to categorize tweets using sentiment analysis.  This entailed changing the data collection, eliminating redundant and little-used data columns, and converting the columns to numerical values. The dataset was then categorized using those numerical values. Dataset preprocessing, feature extraction, and classification are the three processes that make up the entire process.

\subsection{Data Preprocessing}
There are 14 columns in the dataset. One of them provided the classification data. Four other columns offered confidence and gold values, while the remaining provided information about the tweet. But some columns barely contained any information. Fig. 2 illustrates specific columns that have more than 80\% null values.\\
\vspace{-0.1in}
\begin{figure}[ht]
\centerline{\includegraphics[width=0.52\textwidth]{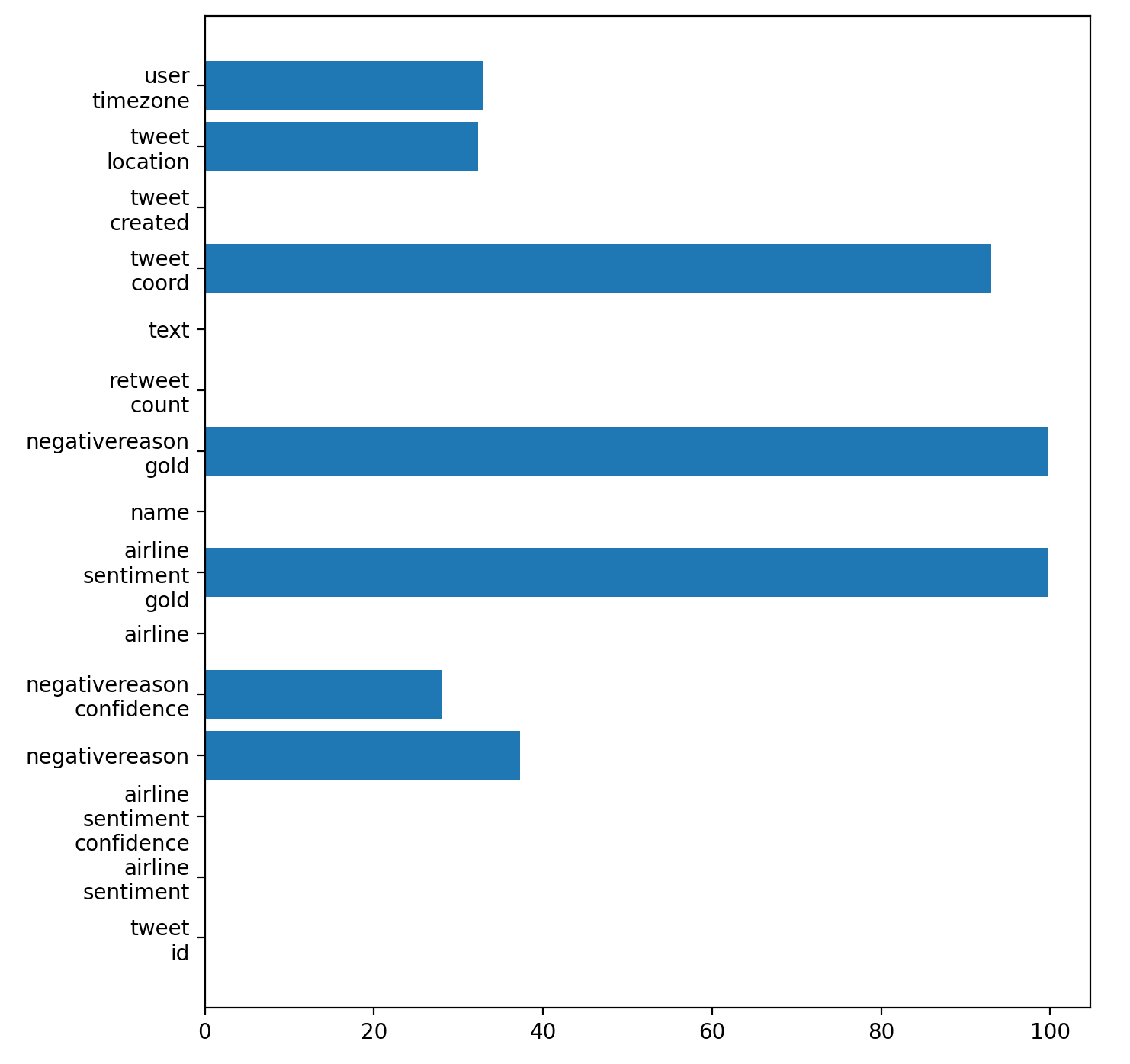}}
\caption{Null value percentile of different columns}
\label{fig2}
\end{figure}
This graph shows that the tweet\_coord, airline\_emotion, and negative\_reason all had more than 80\% null values. Only twitter\_coord was retained out of all these to determine the tweet's coordinates by fusing the details of the tweet's location and the user's time zone. The rest of the columns mentioned were removed. The columns for the Tweet id and confidence value were also removed since they cannot be extracted directly from tweets. As a result, the number of columns dropped from 14 to 9.\\
The tweet text content was in the text column; each tweet text went through 4 stages of the cleaning process and created two new columns to replace the old one. The no\_url text column only filtered the URLs mentioned inside the tweet text. On the other hand, the filtered\_text has filtered URLs, hashtags, usernames, and retweets.\\

\begin{algorithm}[H]
 \caption{Algorithm to Clean Tweet Text Content}
 \begin{algorithmic}[1]
 \renewcommand{\algorithmicrequire}{\textbf{Input:}}
 \renewcommand{\algorithmicensure}{\textbf{Output:}}
 \REQUIRE Text Content of Tweet
 \ENSURE  Text Without URLS , Text Without (URL, Hashtag, Usernames and Retweets)
 \renewcommand{\algorithmicensure}{\textbf{Parameters:}}
 \ENSURE $R_E:$ Python Regular Expression Library, $T_W:$ Tweet Text, $RT:$ Re-Tweet Starting Mark, $R_U:$Regular Expression for Selecting URLs, $T_F:$ Fully Filtered Text, $T_{nu}:$ No Url Text\\
 \vspace{0.1in}
  \textit{Initialisation} : $T_F \leftarrow Null, T_{nu}\leftarrow Null$,\\ $R_U\leftarrow$ r"[(http(s)?):\textbackslash{}/\textbackslash{}/(www\textbackslash{}.)?a-zA-Z0-9@:\%.\_\textbackslash{}+\\\textasciitilde{}\#=]\{2,256\}\textbackslash{}.[a-z\textbackslash{}/A-Z0-9=@:\%\_+.\textasciitilde{}\#?\&]\{2,256\}"\\
  \vspace{0.1in}
  \STATE $T_{nu}$ = $R_E$.sub($R_U$, "", $T_W$) \hfill\textit{//URLs Removed}
  \STATE text = $R_E$.sub("RT @.+","",$T_{nu}$)\hfill\textit{//Retweets Removed}
  \STATE text = $R_E$.sub("@\textbackslash{}w+","",text)\hfill\textit{//Usernames Removed}
  \STATE $T_F$ = $R_E$.sub("\#\textbackslash{}w+","",text)\hfill\textit{//Hashtags Removed}
  \STATE $T_{nu}$ = $T_{nu}$.replace(" \ "," ")
  \STATE $T_F$ = $T_F$.replace(" \ "," ")\hfill\textit{//Double white spaces removed}
  \vspace{0.1in}
 \RETURN $T_F,T{nu}$
 \end{algorithmic}
 \end{algorithm}

\begin{algorithm}[H]
 \caption{Algorithm to prepare dataset for CNN model}
 \begin{algorithmic}[1]
 \renewcommand{\algorithmicrequire}{\textbf{Input:}}
 \renewcommand{\algorithmicensure}{\textbf{Output:}}
 \REQUIRE Prepossessed Data-set
 \ENSURE  Processed Data-set
 \renewcommand{\algorithmicensure}{\textbf{Parameters:}}
 \ENSURE $DATA_{pp}:$ preprossessed data, $DF:$ Processed data, $R_@:$ Remove Username, $R_\#:$ Remove Hashtag, $R_{rt}:$ Remove Re-Tweet, $Q_{aug}:$ Queue for augmanting data, $Q_{class}:$ Queue for preserving text class, $T_{raw}:$ Raw texts, $RE_{List}:$ List for text cleaning methods, $L_{aug}:$ List of augmented data, $NAW:$ nlpung word augmenter method, $NAS:$ nlpung sentence augmenter method, $C_{list}:$ List element text cleaning and duplicate removing method\\
 \vspace{0.1in}
\textit{Initialisation} : $DF \leftarrow Null, RE_{List} \leftarrow [R_@,R_\#,R_{rt}]$
\vspace{0.1in}
\FOR{row in $DATA_{pp}$}
\STATE $T_{raw}$ = text in row
\STATE $Q_{class}$.append(class name in row)
\STATE $Q_{aug}$.append($T_{raw}$)
\FOR{$R_E$ in $RE_{List}$}
\STATE $T_{temp}$ = $R_E$($T_{raw}$)
\IF{$T_{temp}$ != $T_{raw}$}
\STATE $Q_{aug}$.append($T_{temp}$)
\STATE $T_{raw}$ = $T_{temp}$
\ENDIF
\ENDFOR
\ENDFOR
\STATE $L_{aug}$.append($NAW$.augment($Q_{aug}$,$Q_{class}$))
\STATE $L_{aug}$.append($NAS$.augment($Q_{aug}$,$Q_{class}$))
\STATE $DF \leftarrow C_{list}$($L_{aug}$)
\vspace{0.1in}
\RETURN $DF$
\end{algorithmic}
\end{algorithm}

As this work has used SVM, ANN, and CNN models, 14K data might be enough for SVM and ANN models but not for the CNN model. So the data processing at CNN was different. For CNN, more data has been generated using augmentation techniques of nlpung library\cite{b9}. This process includes CPU and GPU run word and sentence level augmenter. word embeddings\cite{b10}, synonym\cite{b11}, back translation(GPU)\cite{b12}, contextual word embeddings for sentence augmentation(GPU)\cite{b13}, abstractive summarization(GPU)\cite{b9} has been used for the augmentation. In the word embedding, the pre-trained glove6B model is used\cite{b14}. Using these resources, \label{Algorithm 2} was used to generate a data set of about 350k samples containing clean text and the airline\_sentiment column.\\

\subsection{Feature Extraction}
The feature extraction process for CNN, SVM, and ANN was slightly different. A common part has been used both for CNN and SVM-ANN. For common cases, the 'airline\_sentiment' column has been converted to a number using the listing method. The text was cleaned and tokenized using the tokenizer library. Features for SVM and ANN have also been extracted from the information the rest of the column provides. The tokenized text list has been converted to float values using a pre-trained model. Glove6B has been used in this work to extract features from the texts. Further, these features have been converted to one feature or seven features using the PCA technique. After that, all the values were normalized.
\subsection{Classification Model Description}
This work has performed CNN on word embedding without any pre-trained models. Here the best performance has been achieved on a CNN model, which has only one hidden layer of 128 neurons—along with this 128 filter of kernel sizes 3, 4, and 6, an embedding layer of 64 dimensions, and a drop layer of 0.5, as shown in Fig. 3.\\
\vspace{-0.1in}
\begin{figure}[ht]
\centerline{\includegraphics[width=0.52\textwidth]{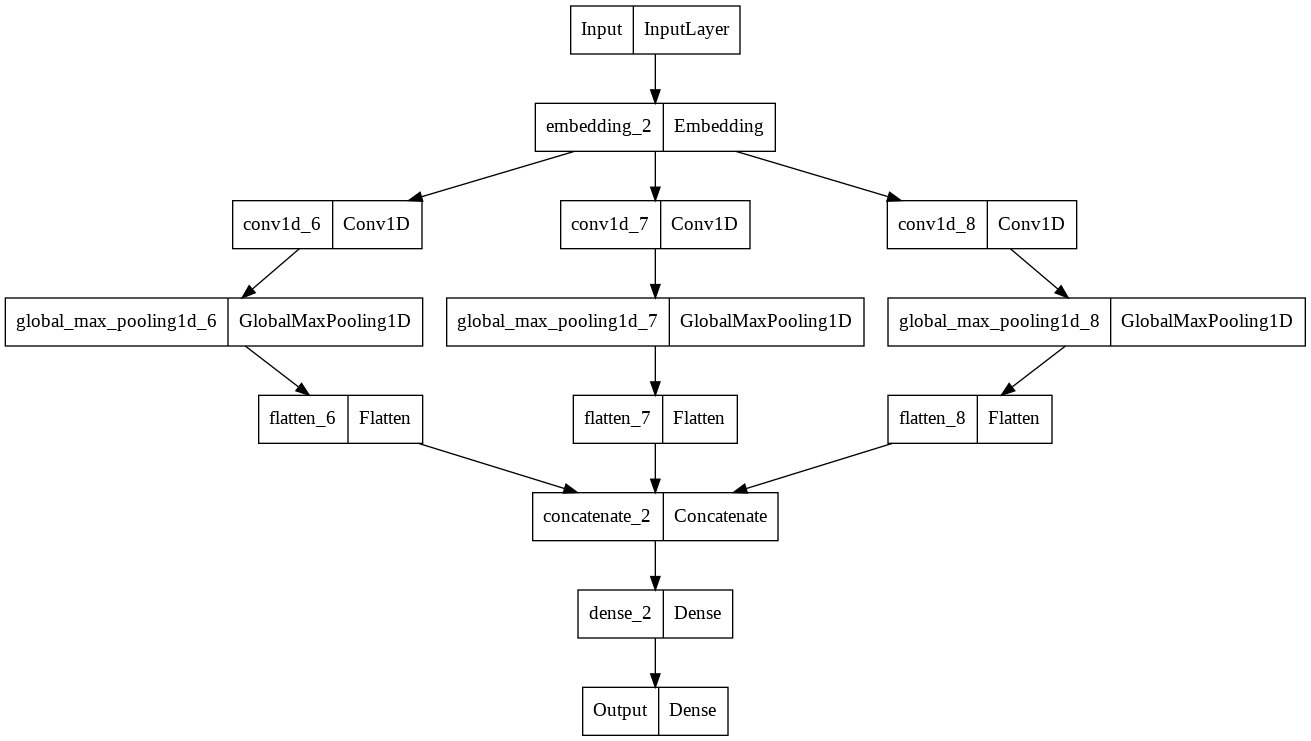}}
\caption{Graphic representation of CNN model}
\label{Fig 3}
\end{figure}

The C-Support Vector Classification algorithm with a C of 10 is utilized for SVM. In comparison, the ANN method has been used in several configurations, listed in Table 1.\\

\begin{tabular}{ |p{1.4cm}|p{1.2cm}|p{1.4cm}|p{1.4cm}|p{1cm}|  }
 \hline
 \multicolumn{5}{|c|}{Table 1: ANN Model Configaration} \\
 \hline
 Configur- ation &Hidden layer&neuron in 1st layer&neuron in 2nd layer&Size of Batch\\
 \hline
 ANN V1 & one & 016 & 000 &256\\
 ANN V2 & one & 032 & 000 &256\\
 ANN V3 & one & 064 & 000 &128\\
 ANN V4 & two & 016 & 004 &256\\
 ANN V5 & two & 032 & 008 &128\\
 ANN V6 & two & 064 & 016 &064\\
 \hline
\end{tabular}

\section{Results and Discussions}
All of the experiments in this study were carried out using Python 3.9 with Keras and the Scikit-Learn package. The experiment was split into two phases: evaluating model performance on the text feature alone, and comparing model performance on all features in a tweet in addition to text. In this case, CNN performed better than the SVM and each of the six ANN configurations, and adding features other than the text of tweets dramatically increased the performance of the models. The accuracy of the CNN model is displayed in Fig. 4.\\
\vspace{-0.3in}
\begin{figure}[ht]
\centerline{\includegraphics[width=0.5\textwidth]{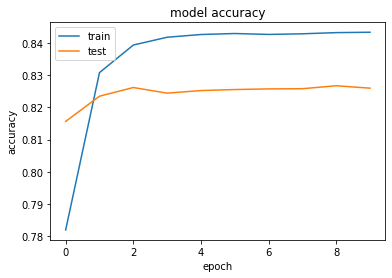}}
\caption{Accuracy result of CNN model}
\label{Fig 4}
\end{figure}

\vspace{-0.1in}
\begin{figure}[ht]
\centerline{\includegraphics[width=0.48\textwidth]{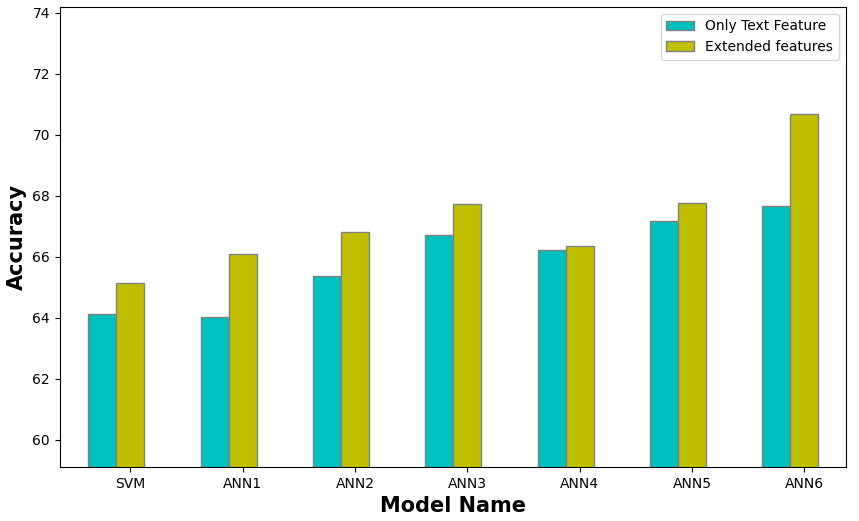}}
\caption{Performance of SVM and ANN models}
\label{Fig 5}
\end{figure}

The accuracy of SVM and ANN models for a distinct set of features is shown in Fig. 5. The "Only text feature" set included the features extracted from the cleaned text of tweets using the glove6B pre-trained model. In contrast, the "extended features" included other features that pulled from the tweet location, time, username, and airline name, along with the features extracted from the texts.\\

\vspace{-0.3in}
\begin{figure}[ht]
\centerline{\includegraphics[width=0.5\textwidth]{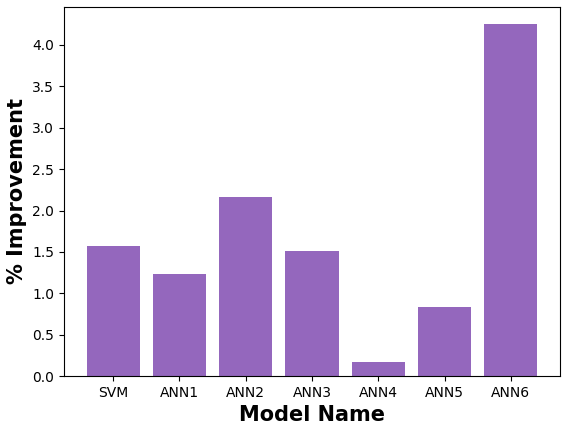}}
\caption{\% improvement of models for extended features}
\label{Fig 6}
\end{figure}
Fig. 6 illustrates how the models have improved as a result of the addition of features taken from sources other than tweet text. And in certain instances, it provides a notable improvement that is greater than 4

\section{Limitations and Future Works}
This work has demonstrated the effect of information other than text that can be extracted from the tweets using SVM and ANN. Due to the small sample size in the initial data set (approximately 14k), it has not been able to perform this in CNN. However, doing the same thing will boost the performance of the CNN model. Therefore, in the future, by creating a data set with sufficient sample size, this can be proved in the CNN model.

\end{document}